\documentclass{IEEEtran}
\usepackage{amsmath,amsfonts}
\usepackage{algorithmic}
\usepackage{algorithm}
\usepackage{array}
\usepackage[caption=false,font=normalsize,labelfont=sf,textfont=sf]{subfig}
\usepackage{textcomp}
\usepackage{stfloats}
\usepackage{url}
\usepackage{verbatim}
\usepackage{graphicx}
\usepackage{cite}
\usepackage{amsmath} % assumes amsmath package installed
\usepackage{amssymb}  % assumes amsmath package installed
\usepackage{amsthm}
\theoremstyle{definition}
\newtheorem{definition}{Definition}[section]
\newtheorem{theorem}{Theorem}[section]

\begin{document}

\title{Encoding Material Safety using Control Barrier Functions for Soft Actuator Control}

\author{Nicholas Pagliocca$^{1,2}$, Behrad Koohbor$^{1,2}$, and Mitja Trkov$^{1,*}$
% <-this % stops a space
\thanks{This work has been submitted to the IEEE for possible publication. Copyright may be transferred without notice, after which this version may no longer be accessible.}
\thanks{This work was partially supported by NSF-CMMI Grant No. 2035660 and NSF Grant No. 2235647.}
\thanks{$^{1}$Department of Mechanical Engineering, Rowan University, Glassboro, NJ, 08028, USA.}%
\thanks{$^{2}$Advanced Materials \& Manufacturing Institute, Rowan University, Glassboro, NJ, 08028, USA.}%
\thanks{$^{*}$Corresponding author: trkov@rowan.edu.}}



\maketitle

\begin{abstract}
Until recently, the concept of soft robot safety was an informal notion, often attributed solely to the fact that soft robots are less likely to damage their operating environment than rigid robots. As the field moves toward feedback control for practical applications, it becomes increasingly important to define what safety means and to characterize how soft robots can become unsafe. The unifying theme of soft robotics is to achieve useful functionality through deformation. Consequently, limitations in constitutive model accuracy and risks of material failure are inherent to all soft robots and pose a key challenge in designing provably safe controllers. This work introduces a formal definition of material safety based on strain energy functions and provides a controller that enforces it. We characterize safe and unsafe sets of an incompressible hyperelastic material and demonstrate that safety can be enforced using a high-order control barrier function (HOCBF) with quadratic program–based feedback control. As a case study, we consider a pressurized hyperelastic tube with inertial effects, first-order viscous effects, and full-state feedback. Simulation results verify that the proposed methodology can enforce the material safety specification.
\end{abstract}


\section{Introduction}
Soft robots manufactured from flexible materials are often assumed to be inherently \textit{safe} and have been shown to be effective in interacting with many delicate environments, having a lower risk of causing damage than rigid systems \cite{RusDaniela2015Dfac, ChenCosimoMazzolai2025Review}. Despite many research studies citing the inherent safety of soft robots, soft robots are in reality far from safe in a physical sense. For example, soft robots and soft actuators can easily damage an object through contact if not regulated properly \cite{OnIntrinsicSafetyofSoftRobots, 10522042, WangHesheng2021HVCo}. Perhaps worse and depending on construction, material failure can occur from rupture, fatigue, cable snapping, electrical fault, and many other sources, all of which can have devastating effects on a soft robot's working environment \cite{OnIntrinsicSafetyofSoftRobots, DorseyKristenL2021LoLM, JeongUseok2021Raoa}.

As the field of soft robotics advances towards control for practical applications, the concept of a formal definition of safety becomes significantly more important and, therefore, is a topic of active research \cite{PattersonSelfContactCBF, XuFan2024HVSCObsAvoidance, dickson2025safe, sue2025reinforcement, sabelhaus2024safeSORO, stolzle2025safe, Sabelhaus2025SensorFault}. Although the aforementioned examples of contact damage and material fault intuitively showcase unsafe soft robot operation in a physical sense, formal definitions of what constitutes unsafe must be developed to prevent their occurrence.

Recent works by Patterson \textit{et al.} \cite{PattersonSelfContactCBF}, Dickson \textit{et al.} \cite{dickson2025safe}, and Wong \textit{et al.} \cite{wong2025contactawaresafetysoftrobots} have posed the question of how to define safety as a technical term for soft robots. In the controls community, safety is typically defined through the satisfaction of constraints and the invariance of a system under dynamics. Barrier functions provide a method ensuring that trajectories that start inside of an invariant set do not reach the complement of the invariant set, which have been extended to controls applications through control barrier functions (CBFs) \cite{CBFOverviewAmes, AmesGrizzleCbfOG}.

Patterson \textit{et al.} examined the use of CBFs and high-order CBFs (HOCBFs) to manage self-contact in a soft-rigid robot \cite{PattersonSelfContactCBF}. Exploring \textit{self-contact safety}, CBFs were used to prevent undesired stiffening and changes in dynamics associated with self-contact. Although soft robots are typically designed to interact with their environments, recent work has also shown that CBF methods can be used for soft robots to avoid obstacles \cite{ZouShuangquan2025MPCfAvoidance} and relax overly restrictive conditions for robots traversing constrained environments, allowing some contact within a safe set of forces \cite{XuFan2024HVSCObsAvoidance}. Since contact is often allowed but not desired in soft robot obstacle avoidance, \textit{collision avoidance safety} relative to soft robot applications can be classified as its own safety category.

Dickson \textit{et al.} formalized the idea of \textit{force safety} and developed a controller using CBFs to guarantee force safety requirements on an end effector in a deformable environment. Similarly, Wong \textit{et al.} showed that HOCBFs combined with high-order control Lyapunov functions can be used to impose contact force limits on the entire body of a soft manipulator while maintaining robot shape \cite{wong2025contactawaresafetysoftrobots}.

The material composition of soft robots is known to offer many challenges in synthesis and actuation, which translate into difficulties in modeling, control, and failure \cite{OnIntrinsicSafetyofSoftRobots, DellaSantinaCosimo2023MCoS, SoftRobotModelingReviewMain, handShearAuxRusCosimo}. A series of recent works have explored the safe supervisory control of shape memory alloy (SMA) actuators, preventing overheating from contact \cite{10521983, jing2022safe}, and rigorously demonstrating a provably safe supervisory controller \cite{sabelhaus2024safeSORO}. Fundamentally, the aforementioned works address the problem of soft robot composition and failure under control, which can be classified as a type of \textit{material safety}.

The current work aims to formalize material safety for soft robotic applications by formulating an invariance condition based on strain-energy functions. A formal concept of material safety and a general method of enforcing it would help mitigate the aforementioned hazards associated with soft robot operation and complement the plethora of safety types in the literature for a more complete concept of safety (\textit{i.e.}, self-contact, collision avoidance, sensor fault, and force safety). The key contributions of this paper include: 

\begin{enumerate}
    \item A formal definition of safety in the sense of material response utilizing strain energy functions,
    \item A HOCBF-QP based control that satisfies the material safety definition for systems with second order dynamics, 
    \item Computational validation of material safety enforcement with a HOCBF-QP based feedback on a hyperelastic tube with inertial and first-order viscous effects in stretch coordinates. 
\end{enumerate}

\section{Theoretical Preliminaries}
Theoretical preliminaries on finite elasticity and HOCBFs are presented to aid the following theoretical developments, analyses, and discussions. %Later, this general theory will be applied to an example of an inflating tube with inertial and viscoelastic effects. 
Boldfaced letters describe tensorial variables. Superscripts are used to classify a variable. Whereas subscripts are associated with a counting index or a coordinate axis, in each case they are identified in writing. The superscript $T$, is reserved for the transpose operation. $diag(\cdot)$ is used to describe a diagonal matrix with zeros in all off-diagonal elements. Explicit calls of time-dependence are removed for brevity unless they are crucial for discussions.

\subsection{Material Modeling}
This section provides a primer on hyperelasticity (Intact Green elastic materials with a reversible path-independent response) and comments on viscoelasticity. 

\subsubsection{Hyperelasticity}
\label{sec:finiteElastic}
Consider a material point with an undeformed configuration, $\mathbf{X}$, and a deformed configuration, $\mathbf{x}$. The mapping relating the configurations through stretches is called the deformation gradient, $\mathbf{F}$, and is expressed in eq. \ref{eq:defGradGeneral}.

\begin{equation}\label{eq:defGradGeneral}
    \textbf{F} = \frac{\partial \mathbf{x} }{\partial \mathbf{X}}
\end{equation}

%where $\nabla$ is the gradient operator. 
The left Cauchy-Green deformation tensor, $\mathbf{B}$, is then calculated from the deformation gradient to describe the stretch as shown in eq. \ref{eq:cauchyGreenDef}.

\begin{equation}
\label{eq:cauchyGreenDef}
\mathbf{B}=\mathbf{F}\mathbf{F}^{T}
\end{equation}

where $\mathbf{B}=\mathbf{B}^{T}$ and $\mathbf{B} \succ 0$. %The choice of left Cauchy-Green tensor was for simplicity in modeling later in the work. 
The invariants of $\mathbf{B}$ in terms of the principal extension ratios $\lambda_{1}$, $\lambda_{2}$, and $\lambda_{3}$ are typical of hyperelastic modeling.
%used in hyperelasticity models (the Ogden model is an example of a model that does not use the invariants, and instead uses stretches directly \cite{OgdenRaymondWilliam1972Ldie}). T
The first invariant of $\mathbf{B}$ is shown in eq. \ref{eq:firstInv} and will be used in later analyses.

\begin{equation}
    \label{eq:firstInv}
    I_{1} = trace(\mathbf{B})
\end{equation}

The measure of the stored energy per unit volume in a material from deformation is characterized by a strain energy function and is commonly used in hyperelastic constitutive models. Strain energy functions are scalar functions that are independent of the strain tensor used in their representation (see eq. 4.3.18 of Ogden \cite{ogden1997non}). In this work, we consider isotropic hyperelastic materials with strain energy functions, $W$, of the form shown in eq. \ref{eq:strainEnergyGeneral}.

\begin{equation}
\label{eq:strainEnergyGeneral}
W(\boldsymbol{\lambda}) = f(\lambda_{1},\lambda_{2},\lambda_{3})
\end{equation}

where $\boldsymbol{\lambda}$ is a stretch vector, dimensions are discussed later. In this work, the Neo-Hookean strain energy function is used for simplicity and is shown in eq. \ref{eq:neoHookean} \cite{RivlinRS1948Ledo}.

\begin{equation}
\label{eq:neoHookean}
W(\boldsymbol{\lambda}) = \frac{\mu}{2}(I_{1}-3)
\end{equation}

where $\mu$ is the low-strain shear modulus of the material and $I_{1}$ is the first principal invariant defined earlier in eq. \ref{eq:firstInv}. In incompressible materials, the product of the stretches must equal one (\textit{i.e.}, $\lambda_{1}\lambda_{2}\lambda_{3} = 1$), thus, at least one stretch is constrained by the other two stretches as shown in eq. \ref{eq:constrainedStretch3}.

\begin{equation}
\label{eq:constrainedStretch3}
    \lambda_{3}= \frac{1}{\lambda_{1}\lambda_{2}}
\end{equation}

The Cauchy stress, $\sigma_{i}$, is used in this work in the principal directions as defined in eq. \ref{eq:cauchyStress}.

\begin{equation}
\label{eq:cauchyStress}
\sigma_{i}(\lambda_{i}) = \lambda_{i}\frac{\partial W(\boldsymbol{\lambda})}{\partial \mathbf{\lambda}_{i}} - p
\end{equation}

where $p$ is a Lagrange multiplier that enforces the incompressibility constraints.

\subsubsection{Viscoelastic Material Response}
\label{sec:viscoGeneral}
Rubbers and many other common materials for soft robotics are known to exhibit viscous behaviors, leading to a combination of time-dependent and time-independent material responses \cite{Youssef_2021Book}. Although there are several visco-hyperelastic models in the literature \cite{YANG2000545, BERGSTROM1998931, UPADHYAY2020103777},
\cite{BergstromBook}, we have not formulated our framework using one of these models for simplicity, and instead introduce dissipative behaviors through a first-order linear damping term that is separate from the hyperelastic term.  %\cite{wong2025contactawaresafetysoftrobots, mustaza2019dynamic}. 

\subsection{Dynamic Modeling}
A generalized dynamic model of an incompressible hyperelastic solid in its principal directions with damping effects and an external forcing function is now presented. A generalized stretch vector is defined by $\boldsymbol{\lambda} \in \mathbb{R}^{n}$. The state $\mathbf{s}= [\boldsymbol{\lambda}, \dot{\boldsymbol{\lambda}}] \in \mathbb{R}^{2n}$. It is assumed that each principal stretch satisfies $\lambda_{i} > 0 \quad \forall t \geq 0$. The symmetric positive definite mass matrix is given by $M(\boldsymbol{\lambda}) \in \mathbb{R}^{n \times n}$. Gravitational and Coriolis effects were assumed to be negligible. First-order damping was incorporated through $\mathbf{F}^{v}(\boldsymbol{\lambda}, \dot{\boldsymbol {\lambda}})$, where it was assumed that the viscoelastic model was independent of the control. The elastic restoring force was defined with $\mathbf{F}^{e}(\boldsymbol{\lambda})\in \mathbb{R}^{n}$. The external input was resolved as the directional forces projected over areas from pressurization, defined by $\mathbf{F}^{ext}(\boldsymbol{\lambda})\mathbf{u}$. It is assumed that all matrices of functions are locally Lipschitz in their arguments. The dynamics of the system are presented in eq. \ref{eq:genSysDyn}.

\begin{equation}
\label{eq:genSysDyn}
\mathbf{M}(\boldsymbol{\lambda}) \ddot{\boldsymbol{\lambda}} + 
\mathbf{F}^{v}(\boldsymbol{\lambda},\dot{\boldsymbol{\lambda}}) + 
\mathbf{F}^{e} (\boldsymbol{\lambda}) = 
\mathbf{F}^{ext}(\boldsymbol{\lambda})\mathbf{u}
\end{equation}

\subsection{Control Barrier Functions}
This section introduces fundamental concepts associated with CBFs and HOCBFs. We assume the dynamics of an affine control system shown in eq. \ref{eq:affineGenDynamics}:

\begin{equation}
\label{eq:affineGenDynamics}
\dot{\mathbf{s}} = f(\mathbf{s})+g(\mathbf{s})\mathbf{u}
\end{equation}

where $\mathbf{s} \in S \subset \mathbb{R}^{n}$ and $\mathbf{u} \in U \subset \mathbb{R}^{q}$ is the control. $f: \mathbb{R}^{n} \rightarrow \mathbb{R}^{n}$, and $g: \mathbb{R}^{q} \rightarrow \mathbb{R}^{n \times q}$ are assumed to be locally Lipschitz continuous functions. $S$ is a closed-state constraint set and $U$ is a closed-control constraint set.

Barrier functions provide methods to ensure that a trajectory starting inside of an invariant set does not reach the complement of the invariant set \cite{CBFOverviewAmes}. For domain, $D$, this safe set, $\mathcal{C}$, is the superlevel set of a smooth function $h: D \subset \mathbb{R}^{n} \rightarrow \mathbb{R}$. Thus, $\mathcal{C}=\{s \in D \subset \mathbb{R}^{n}: h(\mathbf{s}) \geq 0 \}$. Please see \cite {CBFOverviewAmes} for general definitions of forward invariance and safety, and Khalil for formal definitions on Class $\mathcal{K}$ functions and relative degree \cite{Khalil_2002}. Less formally, the relative degree of a smooth function $h(\mathbf{s})$ relative to eq. \ref{eq:affineGenDynamics} is the number of times it must be differentiated until the control input appears in the derivative. HOCBFs are used in our formulations due to the relative degree of our HOCBF candidate in Sec. \ref{sec:MaterAwareCBFForm}, and their general flexibility \cite{HOCBFTrans, HOCBFConf}.

A few additional preliminaries are needed to define a HOCBF. First, a constraint function $h(\mathbf{s}): \mathbb{R}^{n} \rightarrow \mathbb{R}$ of relative degree $m$, such that $h(\mathbf{s}) \geq 0$. Second, a sequence of functions $\psi_{i}: \mathbb{R}^{n} \rightarrow \mathbb{R}, i \in \{ 1, \dots, m\}$ is required as shown in eq. \ref{eq:seqOfFunc}: 

\begin{equation}
\label{eq:seqOfFunc}
\psi_{i}(\mathbf{s}) = \psi_{i-1} +\alpha_{i}(\psi_{i-1}(\mathbf{s})), i \in \{1 \dots m \}
\end{equation}

where $\psi_{0}(\mathbf{s})=h(\mathbf{s})$ and $\alpha_{i}$ is a class $\mathcal{K}$ function. Last, a sequence of safe sets, $\mathcal{C}_{i}$ is defined in eq. \ref{eq:sequenceOfSafeSets}.

\begin{equation}
\label{eq:sequenceOfSafeSets}
\mathcal{C}_{i} = \{\mathbf{s} \in \mathbb{R}^{n}: \psi_{i-1}(\mathbf{s}) \geq 0, i \in \{1 \dots m \}
\end{equation}

\begin{definition}[HOCBF \cite{HOCBFTrans}]
\label{def:HOCBF}
Let $\psi_{i}$, $i \in \{1 \dots m \}$ be defined by eq. \ref{eq:seqOfFunc} and $\mathcal{C}_{i}$, $i \in \{1 \dots m \}$ be defined by eq. \ref{eq:sequenceOfSafeSets}. A function $h(\mathbf{s}): \mathbb{R}^{n} \rightarrow \mathbb{R}$ is a candidate HOCBF of relative degree $m$ for the system defined in eq. \ref{eq:affineGenDynamics} if there exist differentiable class $\mathcal{K}$ functions $\alpha_{i}$, $i \in \{1 \dots m \}$ such that 

\begin{multline}
    \label{eq:HOCBF}
    \underset{\mathbf{u} \in \mathcal{U}}{\sup} \quad 
    L_{f}^{m}h(\mathbf{s}) +
    L_{g}^{m}L_{f}^{m-1}h(\mathbf{s})\mathbf{u} + \\
    O(h(\mathbf{s})) + 
    \alpha_{m}(\psi_{m-1}(\mathbf{s}))
    \geq 0
\end{multline}

for all $\mathbf{s} \in \mathcal{C}_{1} \cap \dots \cap \mathcal{C}_{m}$. Where, $L_{f}$ and $L_{g}$ are the Lie derivatives along $f$ and $g$, respectively. $O(h(\mathbf{s}))$ encapsulates higher order Lie derivatives and possible time-varying terms.

\end{definition}

We refer the reader to Xiao and Belta for additional theoretical details of HOCBFs that are beyond the scope of this article \cite{HOCBFTrans}. If a HOCBF candidate satisfies Definition \ref{def:HOCBF} with the accompanying safe sets, then any Lipschitz continuous controller $\mathbf{u} \in U$ that satisfies the constraint in eq. \ref{eq:HOCBF} renders the intersection of sets $\mathcal{C}_{1} \cap \dots \cap \mathcal{C}_{m}$ forward invariant for the system defined in eq. \ref{eq:affineGenDynamics}.

\section{Material Aware HOCBF Formulation} \label{sec:MaterAwareCBFForm}

This section presents a HOCBF-QP based control that will satisfy a formal definition of material safety characterized by a materials strain energy function. The dynamics presented in eq. \ref{eq:genSysDyn} are first rewritten in control-affine form in \ref{eq:genControlAffineDyn}, where $\mathbf{0}_{n \times q}$ is a matrix of zeros.

\begin{multline}
\label{eq:genControlAffineDyn}
    \dot{\boldsymbol{s}} = 
    \begin{bmatrix}
        \dot{\boldsymbol{\lambda}} \\ 
        -\mathbf{M}(\boldsymbol{\lambda})^{-1}( 
\mathbf{F}^{v}(\boldsymbol{\lambda},\dot{\boldsymbol{\lambda}}) + 
\mathbf{F}^{e} (\boldsymbol{\lambda}))
    \end{bmatrix} \\
    +
    \begin{bmatrix}
        \mathbf{0}_{n \times q} \\
        \mathbf{M}^{-1}(\boldsymbol{\lambda})\mathbf{F}^{ext}(\boldsymbol{\lambda})
    \end{bmatrix}
    \mathbf{u}
\end{multline}

Building on hyperelasticity failure models \cite{VolokhKY2007Hwsf}, CBF theory \cite{CBFOverviewAmes, HOCBFTrans}, and recent advances in soft robot control \cite{PattersonSelfContactCBF, dickson2025safe, wong2025contactawaresafetysoftrobots}, we propose that a formal notion of material safety is well represented by strain energy functions. Since strain energy functions relate any strain/stretch state to a scalar function, they provide a compact and highly generalizable representation of a continuum-based robotic system. More importantly, they are easily incorporated into a continuum model through elastic restoring forces. Material safety is formalized in Definition \ref{def:materSafety}.

\begin{definition}[Material Safety] The state of a soft robot, $\mathbf{s}$, is safe in the sense of material response if the maximal strain-energy for all points in the robot continuum at some instant in time, $W(\boldsymbol{\lambda}(t))$, remains within a safe set of strain energy, $\mathcal{W}^{safe} > 0$, and is invariant under the soft robots dynamics. Formally, $W(\boldsymbol{\lambda}(t_{0})) \in \mathcal{W}^{safe} \implies W(\boldsymbol{\lambda}(t)) \in \mathcal{W}^{safe}$ $\forall t \geq 0$.  
\label{def:materSafety}
\end{definition}

The difference of a critical strain energy and a current admissible strain energy is used to formulate a HOCBF in eq. \ref{eq:candidateCBF}.

\begin{equation}
\label{eq:candidateCBF}
h(\mathbf{s})=\mathcal{W}^{safe} - W(\boldsymbol{\lambda})
\end{equation}

where it is assumed that the smoothness of the strain energy function is of the same order as relative degree the dynamics (\textit{i.e.},  $W(\boldsymbol{\lambda}) \in C^{k}$ s.t. $k \geq m$). Since, $0 \leq W(\boldsymbol{\lambda}) \leq \mathcal{W}^{safe}$, $h(\mathbf{s})\geq 0$ for all stretch combinations in the safe set. The choice of a critical strain energy could be based on failure, a factor of safety applied to failure, or a more complicated function accounting for phenomena such as material softening.
In this work, we consider $\mathcal{W}^{safe}$ as the strain energy at the stretch where a constitutive model found with eq. \ref{eq:neoHookean} diverges from experimental data.

\begin{figure}%[!t]
\centering
\includegraphics[width = 3.0in]{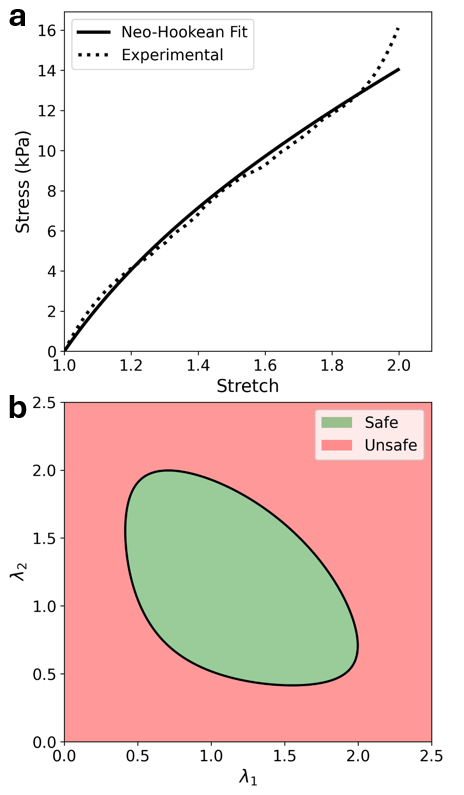}
\caption{(a). Shows the stress-stretch response of a representative Ecoflex 00-50 tensile specimen. (b). Provides a visualization of the safe set of the material.}
\label{fig:materialResp_safeSet}
\end{figure}

Here, we provide a tangible example of what the safe and unsafe sets look like for an incompressible hyperelastic material and a method to find them. Tensile test data from an ASTM D412 Type-A sample from Pagliocca \textit{et al.} \cite{PaglioccaNicholas2023MSRA, ASTM_D412_16_R2021} of Ecoflex 00-50 material (Smooth-On, Inc., Macungie, PA, USA) is provided in Fig. \ref{fig:materialResp_safeSet}a and was used to find the constitutive parameter of the material. 

We considered $\mathcal{W}^{safe}$ as the strain energy at a stretch of 2, since this is where the constitutive model diverged from the experimental data in Fig. \ref{fig:materialResp_safeSet}a. In uniaxial tension, the principal stretches were $\boldsymbol{\lambda}_{uniaxial} = (\lambda_{1},1/\sqrt{\lambda_{1}},1/\sqrt{\lambda_{1}})$, where $\lambda_{1}$ was the stretch in the loading direction. The principal stretches for uniaxial tension were substituted into eq. \ref{eq:neoHookean} to estimate $\mathcal{W}^{safe}$. Using the stretch constraints in eq. \ref{eq:constrainedStretch3} and the Neo-Hookean strain energy function in eq. \ref{eq:neoHookean}, a grid of stretch combinations was computed and fed into the barrier function in eq. \ref{eq:candidateCBF}. The safe set was characterized by stretch combinations that lead to positive $h(\mathbf{s})$, while the unsafe set was the complement of the safe set. These sets are visualized in Fig. \ref{fig:materialResp_safeSet}b. Fortunately, the safe region in Fig. \ref{fig:materialResp_safeSet}b is convex, so there is no immediate need to approximate the safe sets in later formulations.

Moving forward, the next goal is to find the relative degree. The first and second derivatives of the HOCBF are shown in eq. \ref{eq:firstGenLie} and eq. \ref{eq:secondGenLie}, respectively.

\begin{equation}
\label{eq:firstGenLie}
\dot{h}(\mathbf{s})= - \nabla_{\boldsymbol{\lambda}}W(\boldsymbol{\lambda})^{T} \dot{\boldsymbol{\lambda}}
\end{equation}

\begin{equation}
\label{eq:secondGenLie}
\ddot{h}(\mathbf{s}) =  - \dot{\boldsymbol{\lambda}}^{T} \nabla^{2}_{\boldsymbol{\lambda}} W(\boldsymbol{\lambda}) \dot{\boldsymbol{\lambda}}- \nabla_{\boldsymbol{\lambda}} W(\boldsymbol{\lambda})^{T}\ddot{\boldsymbol{\lambda}}
\end{equation}

It is assumed that $\nabla_{\boldsymbol{\lambda}} W(\boldsymbol{\lambda})$ and $\nabla^{2}_{\boldsymbol{\lambda}} W(\boldsymbol{\lambda})$ are locally Lipschitz in their arguments. The dynamics model in eq. \ref{eq:genSysDyn} may be expressed in terms of the second stretch derivative, where the control appears on the right side. From eq. \ref{eq:secondGenLie} it is then visible that the HOCBF has relative degree two. From Definition \ref{def:HOCBF}, and the equations invoked therein, the HOCBF can be used as a constraint for a QP as shown in eq. \ref{eq:QPMain}.

\begin{multline}
    \mathbf{u}(\mathbf{s}) = \underset{u}{\min} \quad \frac{1}{2} ||\mathbf{u}-\mathbf{u}^{nom}||^{2}  \\
    \text{s.t.} \
     \nabla_{\boldsymbol{\lambda}}W(\boldsymbol{\lambda})^{T}\mathbf{M} (\boldsymbol{\lambda})^{-1}\mathbf{F}^{ext}(\boldsymbol{\lambda}) \mathbf{u} \leq\\
    - \dot{\boldsymbol{\lambda}}^{T} \nabla^{2}_{\boldsymbol{\lambda}} W(\boldsymbol{\lambda}) \dot{\boldsymbol{\lambda}} 
    + \nabla_{\boldsymbol{\lambda}} W(\boldsymbol{\lambda})^{T}
    \mathbf{M}(\boldsymbol{\lambda})^{-1}( \\
\mathbf{F}^{v}(\boldsymbol{\lambda},\dot{\boldsymbol{\lambda}}) + 
\mathbf{F}^{e} (\boldsymbol{\lambda})) \\
+\alpha_{2}( - \nabla_{\boldsymbol{\lambda}}W(\boldsymbol{\lambda})^{T} \dot{\boldsymbol{\lambda}}+\alpha_{1}(\mathcal{W}^{safe} - W(\boldsymbol{\lambda}))
 \label{eq:QPMain}
\end{multline}

where $\mathbf{u}^{nom}$ is the nominal control input. A formal theorem for \textit{Material Safety of Systems with $2^{nd}$ Order Dynamics Under HOCBF-Constrained QP Feedback} is presented in Theorem \ref{thm:thm1} for completeness, which mostly follows from the results of \cite{HOCBFTrans} for a specific set of dynamics.

\begin{theorem}[Material Safety of Systems with $2^{nd}$ Order Dynamics Under HOCBF-Constrained QP Feedback]
\label{thm:thm1}
Consider a system governed by the dynamics in eq. \ref{eq:genControlAffineDyn} such that $\mathbf{s} \in \mathcal{C}$, with the HOCBF defined in eq. \ref{eq:candidateCBF}. Any Lipschitz continuous controller $\mathbf{u}(\mathbf{s})$ found from eq. \ref{eq:QPMain} guarantees the set $\mathcal{C}$ is forward invariant, thus satisfying Definition \ref{def:materSafety}.
\end{theorem}

\begin{proof}
    Theorem 4 of Xiao and Belta establishes forward invariance for an HOCBF with the form shown in Definition \ref{def:HOCBF} with the collection of sets defined in eq. \ref{eq:sequenceOfSafeSets}  \cite{HOCBFTrans}. Therefore, assuming a Lipschitz continuous controller under the dynamics in eq. \ref{eq:genControlAffineDyn}, the HOCBF satisfies the condition in Definition \ref{def:materSafety}. \textit{i.e.}, $W(\mathbf{s}(t_{0})) \in \mathcal{C} \implies W(\boldsymbol{\lambda}(t)) \in \mathcal{W}^{safe}$ $\forall t \geq 0$.
    
\end{proof}

\section{Kinematic and Dynamics Model of Hyperelastic Tube with Inertial and Viscous Effects}
\label{sec:tubeModelGen}

\begin{figure}[!t]
\centering
\includegraphics[width=3.5in]{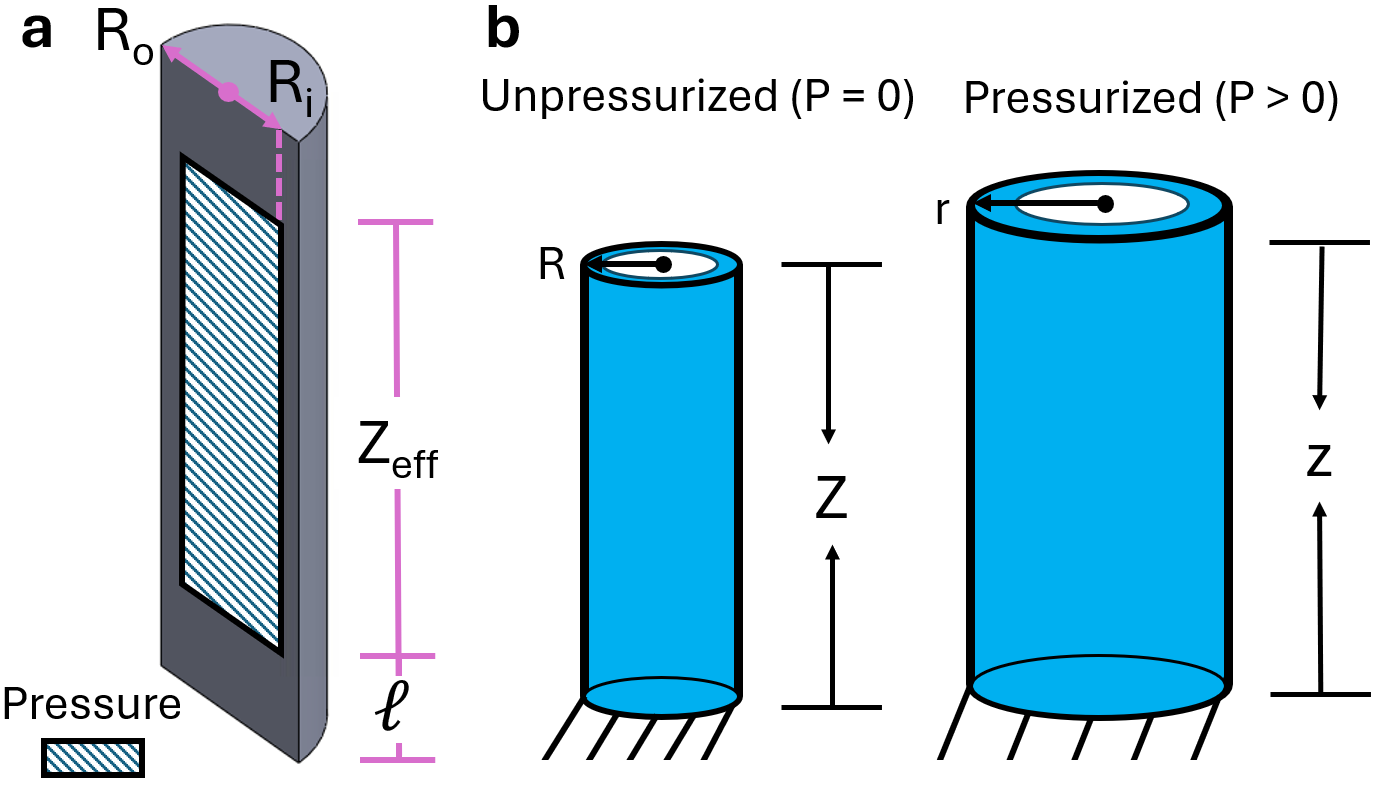}
\caption{(a) Shows an annotated schematic of the tube, The dashed region in the middle of (a) shows where pressure was applied. (b) Shows schematics of an unpressurized (left) and pressurized (right) tube with their principal coordinates annotated.
}
\label{fig:FBD}
\end{figure}

In this section, we provide a simple example showing how to incorporate material safety into a soft robot controller. Our demonstrating example will consider the uniform inflation of a hyperelastic tube with inertial and first-order viscous effects in its principal directions using the general dynamics model from eq. \ref{eq:genSysDyn}. A practical application of these tubes lies in the modeling of the inner layer of soft actuators with fiber \cite{PolygerinosPanagiotis2015MoSF, ConnollyFionnuala2017Adof} or mechanical metamaterial reinforcements \cite{PaglioccaNicholas2023MSRA, WangDong2024MaDo}.
Our tube design was inspired by the work presented in \cite{PaglioccaNicholas2023MSRA} (see Supporting Information therein). The height of the end caps was $\ell$ = 20 mm. The effective length of the tube, $Z_{eff}$ = 90 mm, was described by the height of the unformed tube with the end caps removed. The inner and outer radii of the tube were $R_{i} $ = 10.21 mm and $R_{o}$ = 14.43 mm, respectively. Last, the wall thickness was given by $t$. An annotated model and a schematic of the undeformed and deformed tube configurations are shown in Fig. \ref{fig:FBD}.

In Fig. \ref{fig:FBD}, the deformed coordinates were given by $\mathbf{x}= (r, \theta, z)$ and the undeformed coordinates were given by $\mathbf{X}=(R, \Theta, Z)$. The stretches that took the undeformed coordinates to the deformed coordinates were $\boldsymbol{\lambda}=(\lambda_{r}, \lambda_{\theta}, \lambda_{z})$. The principal stretches were described in the global coordinate frame as $\lambda_{r}$, $\lambda_{\theta}$, and $\lambda_{z}$, corresponding to the radial, circumferential, and axial directions, respectively.  The following assumptions have been made in the modeling:

\begin{enumerate}
    \item Incompressible material response,
    \item Uniform directional stretch in each direction,
    \item The pressure applied is greater than atmospheric pressure,
    \item Forces are resolved via area projection on the axes,
    \item No twisting about the $\Theta$ direction,
    \item Traction-free radial boundary,
    \item Full state access for feedback control.
\end{enumerate}

Since the material was incompressible, the radial stretch was constrained by the axial and circumferential stretches (\textit{i.e.}, $\lambda_{r}= \lambda_{\theta}^{-1}\lambda_{z}^{-1}$). The radial stretch constraint lead to stretches $\boldsymbol{\lambda} \in \mathbb{R}^{2} = [\lambda_{\theta}, \lambda_{z}]$. Applying this constraint, yielded the deformation gradient shown in eq. \ref{eq:defGradCylindrical}.

\begin{equation}\label{eq:defGradCylindrical}
    \textbf{F}(\boldsymbol{\lambda}) = diag(\frac{1}{\lambda_{\theta}\lambda_{z}} , \lambda_{\theta} , \lambda_{z})
\end{equation}

The stresses were resolved using eqs. \ref{eq:defGradCylindrical}, \ref{eq:neoHookean}, and, \ref{eq:cauchyStress}. The constraint in eq. \ref{eq:cauchyStress} was found as $p= \mu \lambda_{r}^{2}$. Since the model was derived in the principal directions, the forces were aligned with the axes and the cross sections remained perpendicular. The elastic restoring forces were as shown in eq. \ref{eq:elasticRestForce}, where $R_{m}=\frac{1}{2}(R_{o}+R_{i})$.

\begin{equation}
\label{eq:elasticRestForce}
\mathbf{F}^{e}(\boldsymbol{\lambda}) =  \mu 
\begin{bmatrix}
    Z_{eff} t%(R_{o}-R_{i})
    (\lambda_\theta^{2}-\frac{1}{\lambda_{\theta }^{2}\lambda_{z}^{2}})
    \quad 2 R_{m}t\pi (\lambda_z^{2}-\frac{1}{\lambda_{\theta }^{2}\lambda_{z}^{2}})
\end{bmatrix}^{T}
\end{equation}

First-order damping was incorporated to account for some of the time-dependent viscous material response as shown in eq. \ref{eq:viscForce}.

\begin{equation}
\label{eq:viscForce}
\mathbf{F}^{v}(\boldsymbol{\lambda}, \dot{\boldsymbol{\lambda}}) = \eta 
    \begin{bmatrix}
        Z_{eff} t \dot{\lambda}_{\theta} &
        2 R_{m}t\pi \dot{\lambda}_{z}
    \end{bmatrix}^{T}
\end{equation}

The mass matrix was derived using the kinetic energy of the continuum for the reference configuration. A change of coordinate matrix was left multiplied to convert the displacement to stretches ($J^{-T} = diag(R_{i},Z_{eff})^{-1}$) to ensure unit consistency with the stretch coordinates, yielding the mass matrix shown in eq. \ref{eq:massMatrix}.

\begin{equation}
    \label{eq:massMatrix}
    \mathbf{M} = 
    diag(
        \frac{\rho \pi(R_{o}^{4}-R_{i}^{4})Z_{eff}}{2 R_{i}},
          \frac{\rho \pi}{3}(R_{o}^{2}-R_{i}^{2})Z_{eff}^{2}
    )
\end{equation}

It is important to note that the units of the mass matrix were $kg \cdot m$ due to the stretch-coordinate formulation of state variables. Virtual work was used to resolve the directional loads from the pressure potential in $\mathbf{F}^{ext}(\boldsymbol{\lambda})$. At some instant in stretch, the volume $V(\lambda_{\theta}, \lambda_{z})=\pi R_{i}^{2}Z_{eff}\lambda_{\theta}^{2}\lambda_{z}$ was differentiated with respect to the stretch states and then mapped to the reference configuration. The external input vector is shown in eq. \ref{eq:extForce}, where $\mathbf{u}$ is the control input from pressure.

\begin{equation}
\label{eq:extForce}
\mathbf{F}^{ext}(\boldsymbol{\lambda})\mathbf{u} = 
\begin{bmatrix}
    2\pi R_{i} \lambda_{\theta} Z_{eff} \lambda_{z} &
    \pi R_{i}^{2} \lambda_{\theta}^{2} 
\end{bmatrix}^{T} \mathbf{u}
\end{equation}

Last, the gradient and Hessian of the strain energy function were required for the constraints in eq. \ref{eq:QPMain}, which are shown in eq. \ref{eq:neoHookeanGradient} and eq. \ref{eq:neoHookeanHessian}, respectively.

\begin{equation}
    \label{eq:neoHookeanGradient}
    \nabla_{\boldsymbol{\lambda}} W(\boldsymbol{\lambda}) = \mu 
    \begin{bmatrix}
        \lambda_{\theta}- \frac{1}{\lambda_{\theta}^{3}\lambda_{z}^{2}} &&
        \lambda_{z}- \frac{1}{\lambda_{\theta}^{2}\lambda_{z}^{3}}
    \end{bmatrix}^{T}
\end{equation}

\begin{equation}
    \label{eq:neoHookeanHessian}
    \nabla^{2}_{\boldsymbol{\lambda}} W(\boldsymbol{\lambda}) = \mu
    \begin{bmatrix}
        1 +  \frac{3}{\lambda_{\theta}^{4}\lambda_{z}^{2}} &
        \frac{2}{\lambda_{\theta}^{3}\lambda_{z}^{3}}\\
        \frac{2}{\lambda_{\theta}^{3}\lambda_{z}^{3}} & 
        1 +  \frac{3}{\lambda_{\theta}^{2}\lambda_{z}^{4}}
    \end{bmatrix}
\end{equation}

\section{Theoretical Model Validation}
\begin{figure}[!t]
\centering
\includegraphics[width=3.5in]{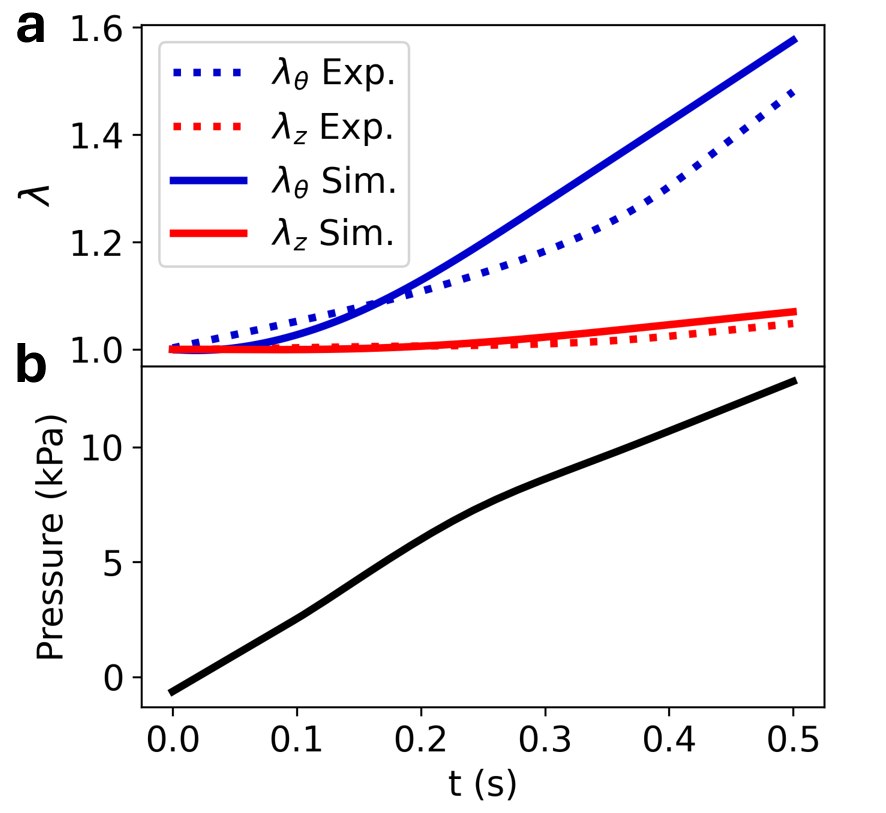}
\caption{Experimental validation of the theoretical model model. (a) Shows the stretch states, while (b) shows the pressure history used in the data set and simulations. 
}
\label{fig:ModelValidation}
\end{figure}

The dynamics of the tube were compared with an existing data set from Pagliocca \textit{et al.} \cite{PaglioccaNicholas2023MSRA}. Using the experimentally acquired pressure history therein, the tube dynamics were simulated forward in time using a 4th order Runge-Kutta integrator and compared to the models stretch states, $\boldsymbol{\lambda}$. 
Direct comparisons of the directional stretches from the theoretical model and the experimental data are shown in Fig. \ref{fig:ModelValidation}a, while the pressure history used in the experimental data set and for the simulations is shown in Fig. \ref{fig:ModelValidation}b. 

Consistent with the experimental data, the results showed that without radial confinement, the radial motions were dominant when compared with the axial motions. The trend and magnitude of the stretches were consistent between the model and the validation data, with larger discrepancies on the $\theta$-axis than on the $z$-axis. We note that a first-order model was used in our derivations, which are known to only capture moderate strains. It is interesting to note that the simulation and experimental data became nearly parallel around 0.35 s, where the error stopped growing. With all of these points, we assumed that our model sufficiency captured the dynamics of the tube for the purposes of demonstrating an application of material safety enforcement in simulation.

\section{Safe Controller Simulation}
Simulations were performed to determine if the controller in eq. \ref{eq:QPMain} satisfied material safety as discussed in Theorem \ref{thm:thm1}. A nominal pressure input with a half-sinusoid profile was selected to test points along the boundary, which was shut off at 0.5 s to allow the dynamics to begin to settle as shown in eq. \ref{eq:nominalControl}.

\begin{multline}
\label{eq:nominalControl}
    u^{nom}(\mathbf{s}) =
\begin{cases}
    10000 sin(2\pi t), & 0 \leq t < 0.5 \\
    0, & t \geq 0.5
\end{cases}
\end{multline}

 The units of pressure in eq. \ref{eq:nominalControl} are Pa. Simulations were performed over 1 second with a $10^{-4}$ timestep. The starting state was $\mathbf{s} = [1,1,0,0]$ (\textit{i.e.}, no pre-stretch and starting at rest). We considered aggressive Class $\mathcal{K}$ parameters $\alpha_{1}$ and $\alpha_{2}$ both equal to 2500. The critical strain energy $\mathcal{W}^{safe}$ = 7.9 kJ/m$^{3}$ was selected as the strain energy at a stretch of 2 from the experimental data in Fig. \ref{fig:materialResp_safeSet}a. This stretch-strain energy pair was selected because the constitutive model diverged from the experimental data in the vicinity of this stretch; thus, any dynamics simulated outside of this range were likely not representative of the material response. The QP in eq. \ref{eq:QPMain} was solved using CVXPY \cite{diamond2016cvxpy, agrawal2018rewriting}.

\begin{figure}[!t]
\centering
\includegraphics[width=3.5in]{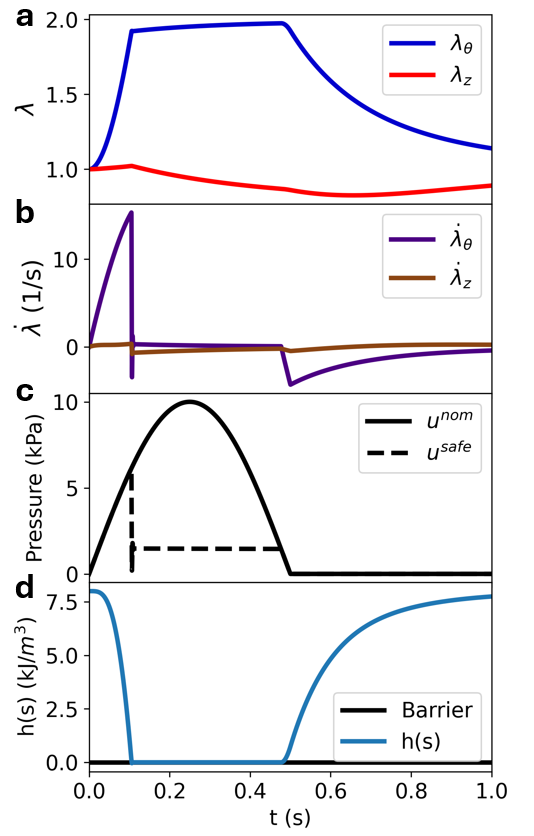}
\caption{Simulation based verification of material safety enforcement with a HOCBF-QP control.
(a) Shows the trajectories of the stretches. (b) Shows the trajectories of the change in stretch as a function of time. (c) Shows the nominal and safe pressure profile. (d) Shows the safety function $h(\mathbf{s})$ evaluated as a function of time, with the safety barrier shown as a line. Since the goal is to verify Theorem \ref{thm:thm1} the nominal trajectories without the safety filter were not shown as they were not relevant.
}
\label{fig:simStudies}
\end{figure}

Fig. \ref{fig:simStudies} presents simulation results for the controller in eq. \ref{eq:QPMain} under the dynamics in eq. \ref{eq:genControlAffineDyn} with the nominal control from eq. \ref{eq:nominalControl}. The stretches are shown in (a), the stretch rates are shown in (b), the nominal ($u^{nom}$) and safe ($u^{safe}$) control inputs are shown in (c), and the evaluation of the HOCBF over the time interval is shown in (d) with the safety barrier shown as a dashed line.

Fig. \ref{fig:simStudies} verifies that Theorem \ref{thm:thm1} enforces material safety since $h(\mathbf{s}) \geq 0$ for the simulation time interval. On the first half of the sinusoid, the stretch and the rate of stretch increased. As the barrier was approached, the controller aggressively dropped the applied pressure while the material continued to respond. As the nominal pressure approached the pseudo-clipped pressure on the downside of the sinusoid, the safe control became inactive and the nominal control was then followed. When the control was turned off the states began to converge to their initial undeformed state, but were still settling at the final simulation time of 1 second.

We note that there was no restriction on negative pressure application, and the states were not constant while the safe controller was active. The large drop in pressure was attributed to the rapid rate of pressurization and inertial effects of the tube, which is discussed in more detail in Sec. \ref{sec:discussions}. The order of magnitude larger changes of $\theta$-axis states than the $z$-axis states, and the decrease in the axial stretch while the safe controller was active were attributed to the fact that the restoring and pressurization forces were much larger on the $\theta$-axis from the model in Sec. \ref{sec:tubeModelGen}

\section{Discussions}
\label{sec:discussions}
Simulation results in Fig. \ref{fig:simStudies} validated Theorem \ref{thm:thm1} showing that a controller can be used to keep stretch combinations within a safe set as depicted in Fig. \ref{fig:materialResp_safeSet}. The tensile test results in Fig. \ref{fig:materialResp_safeSet}a were used to guide the selection of a critical strain energy based on the divergence of the constitutive model and the experimental data. From a modeling and dynamics perspective, it would not make sense use data outside of this range since it would lead to inaccurate dynamics. It should be noted that the apparent material failure of rubber tensile specimens is often over the order of 300\%, but irreversible damage often initiates over finite cycles and far before ultimate fracture \cite{ctx11924996770005201}. This is contrary to the idealized view of hyperelastic materials as path-independent and reversible. However, it is an important detail for material safety and control, since designing the controller solely around ultimate failure would be unrealistic and likely catastrophic.

A first-order model was used to account for some of the time-dependent material response in eq. \ref{eq:viscForce}. It is important to note that the model did not account for constant stress softening or the Mullins effect \cite{BergstromBook}. Examining eq. \ref{eq:candidateCBF} the incorporation of these models would change $\mathcal{W}^{safe}$ to a more complicated damage model, which is left as future work. The gradient and Hessian of eq. \ref{eq:candidateCBF} had cubic terms in the denominators of most elements. Recalling that a low-order Neo-Hookean hyperelasticity model was used in our formulation, the order of these elements will aggressively grow with higher-fidelity constitutive models and may cause numerical instability. Methods for approximating the safe set with convex geometries may remove these high-order terms.

Although our demonstrating example was chosen to be relatively simple and grounded in first principles to convey the idea of material safety for soft robot control, it was limited in the sense that the viscous effects were only first-order, a tightly coupled visco-hyperelastic formulation was not used, reference area projections were used for force mappings, and the model was derived in the principal directions. Nevertheless, the dynamics were consistent with the existing literature, as shown in Fig. \ref{fig:ModelValidation}, and the derived model provided insight into what elements are needed to incorporate material safety into a controller.

The simulation results in Fig. \ref{fig:simStudies} confirmed that the proposed controller successfully maintained trajectories within the material safe set. A high amplitude input was selected to try to push the actuator into the unsafe region as fast and aggressively as possible. The time delay and the large pressure drop from 6 kPa to 1.5 kPa from the safe controller in Fig. \ref{fig:simStudies}d were associated with the inertial and viscous dynamics of the tube. Very large Class $\mathcal{K}$ coefficients were selected for two reasons. First, if material safety is enforced in a practical robot controller, it likely means that there is another control objective that is not possible. However, preventing robot material failure or constitutive model breakdown should take priority over any other objectives for many safety reasons. Second, the QP formulation in eq. \ref{eq:QPMain} optimizes over a single step, so it is imperative to have the controller respond as fast as possible, future advancements should consider a finite horizon.

\begin{figure}[!t]
\centering
\includegraphics[width=3.5in]{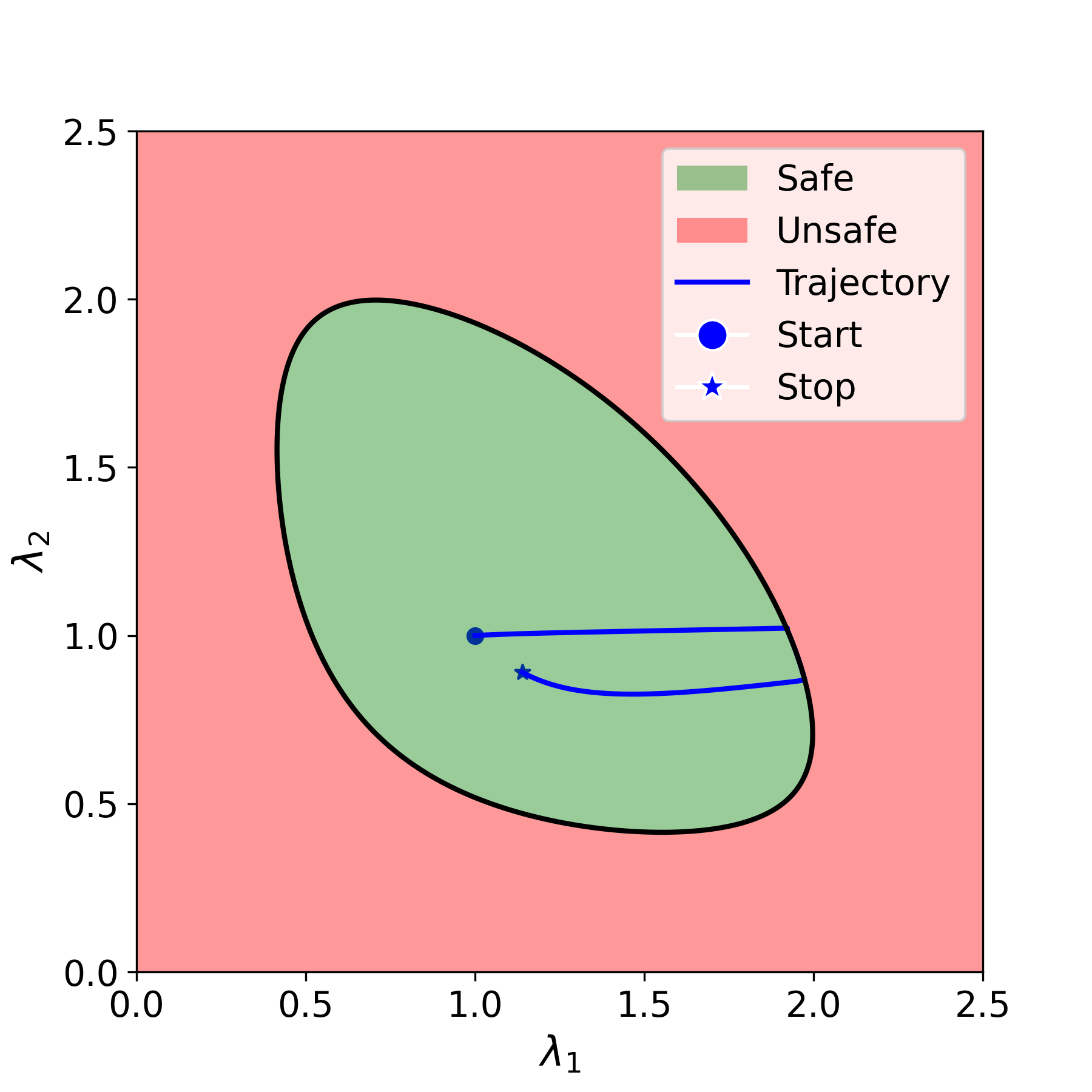}
\caption{Visualization of the safe set with the principal stretches experienced in the tube. The trajectory presents the evolution of the tube stretches according to the safe pressure profile in Fig. \ref{fig:simStudies}c. The circular point marks the start and the star marks the end of the trajectory.
}
\label{fig:safeTraj}
\end{figure}

The near-constant pressure that the safe controller settled to was associated with the pressure required to keep the actuator on the boundary of the safe set. It is important to note that the stretches in Fig. \ref{fig:simStudies}a were changing while $h(\mathbf{s})$ was on the barrier in Fig. \ref{fig:simStudies}d. This means that the stretch pairs were sweeping the edge of the barrier set as shown in Fig. \ref{fig:safeTraj}. The pressure drop arising from the tube dynamics and the state variation when the safe control was active shows that it is likely not sufficient to enforce material safety based on air input/ volume change or a static stretch threshold, and a more robust dynamics-aware formulation is required to enforce Definition \ref{def:materSafety}.

\section{Conclusions}

This work proposed a formal definition of material safety based on strain energy functions and demonstrated its enforcement using a HOCBF-QP controller on a hyperelastic tube with inertial and first-order viscous effects. By addressing issues of material failure and model divergence, our work complements existing safety frameworks in soft robotics. Although material aware safe controllers are of particular use in soft robotics, the authors envision that the methodology developed here can be used to enforce material safety in systems with linear elastic materials where material fault is possible and preventable from control. Future directions include extensions to finite element models, reducing the relative degree for simpler controllers, incorporating advanced visco-hyperelastic and damage models, and exploring integration with MPC and other forms of provably safe control.

\bibliographystyle{ieeetr}
\bibliography{references}

\vfill

\end{document}